%% file: main.tex
\documentclass[journal]{IEEEtran}
\usepackage[T1]{fontenc}
\usepackage{amsmath,amsfonts,amssymb}
\usepackage{algorithm}
\usepackage{algorithmic}
\usepackage{array}
\usepackage{booktabs}
\usepackage{graphicx}
\usepackage{textcomp}
\usepackage{url}
\usepackage{balance}
\usepackage{cite}
\usepackage[hidelinks]{hyperref}
\graphicspath{{figures/}}
\newcommand{\ind}[1]{\mathbf{1}\!\left[#1\right]}
\newcommand{\C}{\mathcal{C}}
\newcommand{\off}{\mathrm{off}}

\newtheorem{rem}{Remark}

\makeatletter
\newcommand{\FirstPageOverview}{%
\begin{minipage}{\textwidth}
\centering
\includegraphics[width=\textwidth]{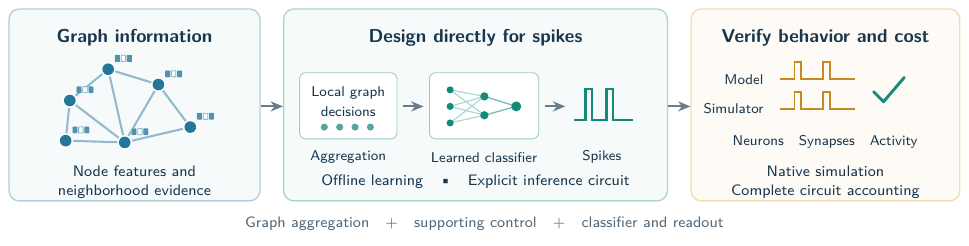}
\def\@captype{figure}
\caption{From graph learning to an explicit spiking realization. Event-driven graph inference requires aggregation, control and readout to be represented in the implementation. DeNA-SGN designs these operations together, learns the classifier offline, and verifies model--circuit agreement in native simulation with complete neuron, synapse and activity accounting. Offline feature encoding and host current decoding are identified separately.}\label{fig:overview}
\end{minipage}\vspace{0.6em}}
\makeatother
\IEEEaftertitletext{\FirstPageOverview}

\input{tables/validation_numbers.tex}
\input{tables/research_findings_numbers.tex}
\begin{document}
    

    
\title{DeNA-SGN: Deterministic Neighborhood Aggregation for Verified Spiking Graph Inference}
\author{Sanjay~Das, Xi~Zhang, Emily~Herron, Prasanna~Date\\
Oak Ridge National Laboratory
\thanks{The authors are with Oak Ridge National Laboratory, Oak Ridge, TN, USA.}}
\markboth{DeNA-SGN --- Research Manuscript Draft}{DeNA-SGN}
\maketitle

\begin{abstract}
Graph learning on neuromorphic substrates requires an explicit spiking realization of neighborhood aggregation, supporting control and classification. We propose DeNA-SGN, which combines fixed OR and tie-inclusive Majority aggregation of binary node features with a learned signed-weight threshold classifier. A degree-dependent inhibitory construction implements Majority through shared control neurons, enabling a complete circuit compilation into SuperNeuroMAT. Across seven datasets and \AllMappedModels{} mapped classifiers, native simulation matches the exported ANN on every evaluated class decision and output-spike pattern.
 On the new confirmatory Amazon Photo dataset, DeNA-SGN achieves \PhotoMean$\pm$\PhotoSD\% current-readout accuracy and \PhotoStrictMean$\pm$\PhotoStrictSD\% strict output-spike accuracy; Wiki-CS reaches \WikiPublicMean$\pm$\WikiPublicSD\% over 20 public masks. In the matched Cora study, we achieve 74.90$\pm$ 2.67\% on current-readout and  68.57$\pm$2.94\% strict spike accuracy. Complete circuit and activity accounting supports analytical energy comparisons and identifies the cost of clocking inactive neurons in large deployments. Current argmax is decoded on the host, and energy is modeled rather than measured on hardware. These findings establish a reproducible route to useful graph classification with decision-matched spiking execution and explicit implementation costs.
\end{abstract}
\begin{IEEEkeywords}
Spiking neural networks, graph neural networks, neuromorphic computing, node classification, ANN-to-SNN mapping, threshold networks, straight-through estimator.
\end{IEEEkeywords}

\section{Introduction}
Graph neural networks (GNNs) learn from node attributes and relations, making them useful for citation analysis, recommendation and other structured prediction tasks. Graph attention learns how strongly neighboring information should contribute~\cite{velickovic2018gat}, while neuromorphic systems organize computation around discrete events and local state updates~\cite{roy2019spike}. Connecting these approaches requires more than substituting a spiking classifier for a conventional output layer. The neighborhood operator, synchronization, control and readout must have explicit implementations whose behavior and cost can be evaluated. Figure~\ref{fig:overview} presents the design objective of this work.

Existing spiking graph methods provide complementary starting points. SpikingGCN constructs a real-valued graph representation before stochastic spike encoding~\cite{zhu2022spikinggcn}. Graph-as-network methods instead realize citation relations as synapses and use online spike-timing-dependent plasticity (STDP), including the NEST implementation of Cong \emph{et al.}~\cite{cong2022semisupervised}, its Bayesian-optimization and feature-inclusion extension~\cite{cong2023hpofeat}, and the fixed-precision Lava implementation of Snyder \emph{et al.}~\cite{snyder2024transductive}. These methods establish the feasibility of spike-based graph processing. They also motivate a complementary question: can a feature-aware graph classifier be designed from threshold operations so that its complete inference circuit can be constructed, its decisions checked against the trained model, and every supporting neuron and synapse included in the accounting?

\begin{figure*}[!t]\centering\includegraphics[width=\textwidth]{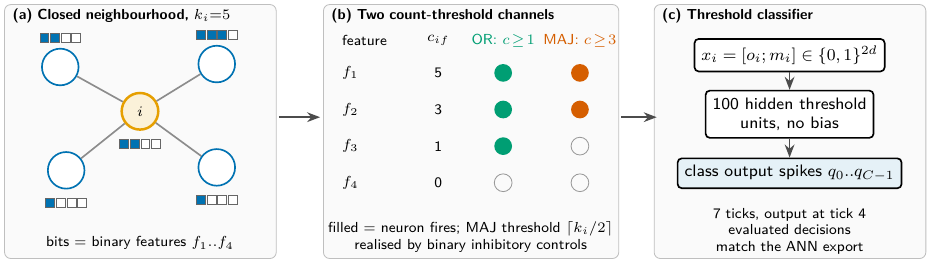}
\caption{Overview of DeNA-SGN. (a) A query node and its closed neighborhood hold binary features. (b) For each feature, an OR neuron fires if any member has it, and a Majority neuron fires if at least $\lceil k_i/2\rceil$ do; the degree-dependent threshold is supplied by shared binary inhibitory control neurons. (c) The binary channels feed a bias-free threshold classifier, and the evaluated circuit decisions match its ANN export.}\label{fig:teaser}\end{figure*}

This paper introduces DeNA-SGN, a spiking graph network that replaces learned neighborhood weighting with fixed logical operators. For every binary feature, an OR neuron indicates whether any node in the closed neighborhood has the feature, and a Majority neuron indicates whether at least half of them do. The resulting binary channels drive a compact threshold classifier. We call the design deterministic in two respects: the aggregation is a fixed function of the graph and its features, with no learned edge weights, and spike generation involves no stochastic encoding, because each query is a single input spike. Fig.~\ref{fig:teaser} gives an overview. The design uses fixed neighborhood summaries rather than learned edge coefficients. This choice exposes the aggregation and classifier as a compact threshold circuit with a precise compilation rule.

The work is guided by four requirements. The aggregation and classifier circuit should be \emph{spike-native}, using threshold neurons and signed synapses without softmax or host-computed gates; offline feature encoding and host current decoding remain outside this circuit. It should be \emph{deterministic}, as described above. It should be \emph{exactly verifiable}, so that exported-model decisions can be checked against native simulation. And it should be \emph{completely costed}, so that every supporting neuron is counted. The main contributions are the following.
\begin{enumerate}
\item \emph{A deterministic, algorithmic aggregation scheme.} DeNA-SGN deploys two fixed logical operators, OR and Majority, applied over each node's closed neighborhood. Its key algorithmic element is a degree-adaptive Majority threshold, $c\ge\lceil k/2\rceil$, implemented by at most $\lfloor\log_2\Delta\rfloor$ shared control neurons with binary-weighted inhibitory synapses. The scheme needs no learned edge weights, no softmax and no stochastic spike encoding (Sections~\ref{sec:method} and~\ref{sec:map}).
\item \emph{Full circuit-level neuromorphic mapping and validation.} We map the complete inference path, including query, relay, control, feature, hidden, output and observer neurons, onto the SuperNeuroMAT simulator. 
\item \emph{Useful predictive performance under validation-only selection.} DeNA-SGN reaches \PhotoMean$\pm$\PhotoSD\% current-readout accuracy on the new Amazon Photo confirmatory dataset and \WikiPublicMean$\pm$\WikiPublicSD\% on Wiki-CS. 
\item \emph{Complete resource accounting and an evidence-based design record.} We count every neuron, synapse, spike and synaptic event, including the supporting relays, controls and observers, and give an analytical energy model with sensitivity bounds. 
\end{enumerate}
The remainder of the paper is organized as follows. Section~\ref{sec:bg} gives background, Section~\ref{sec:rw} reviews related work, Section~\ref{sec:method} presents the method, Sections~\ref{sec:setup} and~\ref{sec:results} describe the experiments and results, Section~\ref{sec:disc} discusses limitations, and Section~\ref{sec:concl} concludes. 

\section{Background}\label{sec:bg}
\subsection{Transductive Node Classification}
A graph with $N$ nodes carries a binary feature matrix $X\in\{0,1\}^{N\times d}$ and labels $y_i\in\{0,\ldots,C-1\}$, of which only a training subset is known. In the transductive setting, all features and edges are visible during training and only training labels enter the loss. The Planetoid versions of Cora, CiteSeer and PubMed~\cite{sen2008collective,namata2012query,yang2016planetoid} are the standard benchmarks; Wiki-CS~\cite{mernyei2020wikics}, ogbn-arxiv~\cite{hu2020ogb} and Zachary's karate club~\cite{zachary1977karate} extend the range of size, class count and difficulty. Cora and CiteSeer features are already binary bag-of-words indicators, which makes them a natural fit for logical aggregation; denser, continuous features must be binarized offline, as described in Section~\ref{sec:setup}.

\subsection{The SuperNeuroMAT Execution Model}
SuperNeuroMAT is a matrix-based simulator of leaky integrate-and-fire neurons with dense and sparse execution~\cite{date2023superneuro,date2026superneuromat,superneuromatdocs}. Our design relies on the following semantics, each confirmed by saved probes: synaptic arrivals are summed with sign before thresholding; a neuron fires when its input strictly exceeds its threshold; with infinite leak, no state survives to the next tick; reset is to zero; a seven-tick refractory period allows each neuron to fire at most once per query; and every synapse has an explicit one-tick delay. STDP is disabled throughout. An artificial-network test $a\ge t$ is therefore realized as the native test $a>\mathrm{nextafter}(t,-\infty)$, which agrees for finite float64 accumulated values and positive finite representable thresholds. This adapter does not guarantee identical rounded sums across different accumulation orders.

\subsection{Training Binary Threshold Networks}
A hard threshold has zero gradient almost everywhere. The straight-through estimator (STE)~\cite{bengio2013ste} keeps the hard threshold in the forward pass and substitutes an identity gradient in the backward pass; it is the simplest member of the surrogate-gradient family~\cite{neftci2019surrogate}. We adopt it because hidden and output firing decisions remain binary in the forward pass, enabling a logical threshold-circuit construction; finite-precision agreement is checked separately.

\subsection{Neighborhood Aggregation as a Thresholded Mean}
Graph convolution in its simplest form replaces a node's features by the mean over its closed neighborhood. For binary features this mean is $c_{if}/k_i$ in the notation of Section~\ref{sec:method}. Binarizing it at two operating points yields the aggregation used here, a link that is made precise in Remark~\ref{rem:mean} and that explains why the operator is a meaningful graph filter even though it has no learned weights.

\section{Related Work}\label{sec:rw}
\subsection{Graph Learning and Spiking Representations} GCN~\cite{kipf2017gcn} averages neighbor features with fixed weights, while GAT~\cite{velickovic2018gat} learns softmax-normalized edge weights; GAT serves as our ANN reference. Both use real-valued activations throughout.

\paragraph{Spiking graph learners} SpikingGCN computes a normalized convolution on real-valued node features, encodes the result as Bernoulli spikes over $T$ steps and classifies with a fully connected spiking layer~\cite{zhu2022spikinggcn}. Its neuromorphic energy analysis is analytical; the authors exclude the signed-spike refinement, SpikingGCN-N, from that hardware-cost analysis. Related work addresses contrastive learning with 1-bit spikes~\cite{li2023spikegcl}, dynamic graphs~\cite{yin2024dsgnn}, continuous-depth propagation~\cite{yin2024cosgnn} and spiking graph transformers~\cite{sun2024spikegraphormer}. Their learning procedures differ: Dy-SIGN uses implicit differentiation to reduce the need to reverse temporal computation~\cite{yin2024dsgnn}. DeNA-SGN instead trains an acyclic threshold classifier offline and verifies a fixed feed-forward inference circuit.

\subsection{Graph-as-Network Neuromorphic Implementations} Cong \emph{et al.} create one neuron per paper plus topic neurons, connect papers by citations and let STDP~\cite{gerstner1996stdp} propagate labels in the NEST simulator~\cite{gewaltig2007nest}, without node features~\cite{cong2022semisupervised}. A follow-up adds Bayesian hyperparameter optimization and studies feature inclusion~\cite{cong2023hpofeat}. Snyder \emph{et al.} give the structure-only network a fixed-precision, Loihi~2-compatible implementation in Lava (Loihi~\cite{davies2018loihi} is the earlier chip)~\cite{snyder2024transductive}. Zhu \emph{et al.} introduce reduced CiteSeer-derived datasets with SuperNeuroMAT baselines~\cite{zhu2026citeseer}, and the simulator paper reports full-graph structure-only accuracies~\cite{date2026superneuromat}. In this family the graph is the network, so inference involves a dynamic process in which plasticity modifies synapses during each query, and the label is read from the learned topic synapses rather than from output spikes.

\begin{table}[!t]\centering\footnotesize\setlength{\tabcolsep}{3pt}
\caption{Design Contrast With the Closest Spiking Graph Classifiers. S-GNN Lineage Refers to~\cite{cong2022semisupervised,cong2023hpofeat,snyder2024transductive}; Its Step Counts Are Those Explored in~\cite{snyder2024transductive}.}\label{tab:design}
\begin{tabular}{@{}>{\raggedright\arraybackslash}p{1.65cm}>{\raggedright\arraybackslash}p{1.3cm}>{\raggedright\arraybackslash}p{1.0cm}>{\raggedright\arraybackslash}p{1.4cm}>{\raggedright\arraybackslash}p{1.3cm}@{}}\toprule
 & Graph operator in spikes & Node features & Learning & Time steps \\\midrule
SpikingGCN & no (real conv.) & yes & offline training & $T$ (tuned) \\
S-GNN lineage & yes (citation synapses) & no / optional & online STDP, BO & 8, 14, 20 \\
\textbf{DeNA-SGN} & yes (OR and Majority) & yes & offline STE, no STDP & 7 (output at 4) \\\bottomrule
\end{tabular}\end{table}
Table~\ref{tab:design} summarizes how DeNA-SGN differs: it uses node features, learns its weights offline, runs no online STDP, and reaches the output at simulator tick 4. The 8 to 20 steps explored in the cited S-GNN study describe different dynamics and do not establish a cross-platform latency advantage.

\subsection{Circuit Verification and Resource Accounting} Simulator comparisons~\cite{kulkarni2021benchmarking} report speed and memory rather than equivalence to a source network. Energy studies of spiking GNNs commonly count multiply-accumulate and synaptic operations at fixed per-operation costs~\cite{zhu2022spikinggcn}; we follow that convention and add sensitivity bounds. Our evaluation couples per-query comparison against a defined ANN export with a constructive full circuit, resource counts and archived execution traces. This makes model behavior, implementation fidelity and assumed operation costs directly inspectable.

\section{The DeNA-SGN Framework}\label{sec:method}
\subsection{Problem Definition and Proposed Pipeline}
We consider transductive node classification on a fixed graph with $N$ nodes, raw features $F\in\mathbb R^{N\times d}$ and labels $y_i\in\{0,\ldots,C-1\}$. Training, validation and evaluation node sets $\mathcal T$, $\mathcal V$ and $\mathcal E$ are disjoint. All node features and graph edges are available as context, but only $y_{\mathcal T}$ enters the differentiable loss. Validation labels select checkpoints, optimization settings and output thresholds; evaluation labels are withheld from these decisions until the selected models are locked.

\emph{Proposal and rationale.} DeNA-SGN replaces learned edge attention with two fixed summaries of binary neighborhood evidence: whether a feature occurs anywhere, and whether at least half of the neighborhood supports it. A learned signed-weight threshold classifier combines these channels. The aim is an explicit threshold-circuit implementation whose decisions and supporting resources can be checked, rather than a claim of lossless aggregation or improved attention. The stages are: (1) encode raw features offline; (2) compute the two channels for training; (3) train and select the classifier; (4) calibrate output thresholds; (5) compile graph, features and learned weights into a static SNN; and (6) simulate a single query spike and decode its outputs. Training uses precomputed aggregate features, whereas native verification computes the aggregation inside the SNN.

\subsection{Binary Feature Encoding and Graph Convention}
Define the binary matrix $X$ and closed-neighborhood adjacency $A$ by
\begin{align}
 X_{jf}&=\ind{F_{jf}>\tau_f},\qquad A_{ij}=\ind{j\in\C(i)},\label{eq:encode}\\
 \C(i)&=\{i\}\cup\{j:(i,j)\in E\ \text{or}\ (j,i)\in E\}.\notag
\end{align}
Duplicate edges are removed. Thus $A$ is symmetric, includes self, and has $k_i=\sum_j A_{ij}\ge1$. Cora, CiteSeer, PubMed and Amazon Photo use $\tau_f=0$ (positive presence), discarding PubMed's nonzero magnitudes. KarateClub uses the 34-dimensional identity matrix and four supplied community labels. Wiki-CS and ogbn-arxiv use the lower median of feature $f$ over $\mathcal T$ only, with equality mapped to zero. These fixed encodings are saved. Encoding and graph construction occur offline; their implementation is outside the deployed circuit and inference-energy model. At compilation, $X$ determines fixed feature-delivery synapses; inference does not inject host-computed aggregate bits.

\subsection{Two-Channel Neighborhood Operator}
Let $b_{jf}=X_{jf}$. The count matrix is $AX$, and the two channels for a query $i$ are
\begin{align}
 c_{if}&=(AX)_{if}=\sum_{j\in\C(i)} b_{jf},\label{eq:count}\\
 o_{if}&=\ind{c_{if}\ge1},\qquad
 m_{if}=\ind{c_{if}\ge\lceil k_i/2\rceil},\label{eq:om}\\
 x_i&=[o_i;m_i]\in\{0,1\}^{2d}.\label{eq:x}
\end{align}
The operator is permutation-invariant and has no learned edge coefficients. OR captures any positive evidence; Majority adds a higher support requirement, with even-size ties included. The pair separates absent features, features below half support, and features meeting half support, but still discards exact counts, neighbor identity and raw magnitude. The learned classifier can weight these coarse regimes differently without computing an edge softmax or multiplying spike amplitudes by input-dependent attention coefficients. This is fixed graph aggregation, not standard GAT.
\begin{rem}\label{rem:mean}
For binary inputs, OR and Majority are thresholds on the unweighted closed-neighborhood mean: $o_{if}=\ind{c_{if}/k_i>0}$ and $m_{if}=\ind{c_{if}/k_i\ge1/2}$. This is a thresholded mean operator, not an identity with the full normalized GCN layer. The same construction can express other count thresholds, although the present validation-selected study fixes the two channels in Eq.~\eqref{eq:om}.
\end{rem}
For $k_i\le2$ the channels coincide; an isolated node reduces to its own features. For $k_i\ge3$, a solitary positive does not activate Majority. A single feature change can nevertheless flip Majority near threshold, so no general single-neighbor robustness guarantee is claimed.

\subsection{Learned Threshold Classifier and Decoders}
Let $\phi=(W_1,W_2)$, with $W_1\in\mathbb R^{H\times2d}$, $W_2\in\mathbb R^{C\times H}$ and $H=100$. Both layers are bias-free. With hidden threshold $\theta_h=1$ and class thresholds $\theta_o\in\mathbb R^C$, initialized to one,
\begin{align}
 a_i&=W_1x_i,\quad h_i=\ind{a_i\ge\theta_h},\notag\\
 z_i&=W_2h_i,\qquad q_i=\ind{z_i\ge\theta_o}.\label{eq:net}
\end{align}
Indicators act componentwise. $W_1$ and $W_2$ are learned signed weights; aggregation coefficients and hidden thresholds are fixed. The OR block copies a Xavier-initialized reference classifier, the Majority block starts at zero, and $W_2$ copies the reference readout. OR-only and OR+Majority therefore start from the same function in the matched study.

The two reported decoding rules answer different questions:
\begin{align}
 \hat y_i^{\rm cur}&=\min\arg\max_c z_{ic},\label{eq:pred}\\
 s_i^{\rm strict}&=\ind{q_i=e_{y_i}},\label{eq:strict}
\end{align}
where $e_{y_i}$ is a class one-hot vector. Current argmax is computed on the host from simulated observer currents. Strict correctness requires exactly the correct class neuron to fire; silence and multiple spikes fail. Because $\theta_o$ is class-specific, strict correctness does not imply current-argmax correctness. Calibrating $\theta_o$ changes $q_i$ but not $z_i$ or $\hat y_i^{\rm cur}$.

\subsection{Training Objective and Recorded Model Selection}
For a training batch $\mathcal B\subseteq\mathcal T$, the implemented objective is
\begin{align}
 \mathcal L_{\mathcal B}
 &=\frac1{|\mathcal B|}\sum_{i\in\mathcal B}\mathrm{CE}(z_i,y_i)
   +\mathcal L_{\rm margin}+0.1\mathcal L_{\rm emb},\label{eq:loss}\\
 \mathcal L_{\rm margin}
 &=\frac1{|\mathcal B|}\sum_i[1.25-z_{i,y_i}]_+\notag\\
 &\quad+\frac1{|\mathcal B|(C-1)}\sum_i\sum_{c\ne y_i}[z_{ic}-0.75]_+,\notag
\end{align}
where $[v]_+=\max(v,0)$ and sums over $i$ in this subsection run over $\mathcal B$. The margins encourage separation around the initial unit firing threshold. For the embedding regularizer, let $u_i=a_i/\max(\|a_i\|_2,10^{-8})$, $P=\{(i,j)\in\mathcal B^2:y_i=y_j\}$ (including self pairs), $Q=\{(i,j)\in\mathcal B^2:y_i\ne y_j\}$ and $\bar p_r=|\mathcal B|^{-1}\sum_i\sigma(a_{ir}-1)$. Then
\begin{align}
 \mathcal L_{\rm emb}
 &=\frac{\sum_{(i,j)\in P}(1-u_i^\top u_j)}{\max(1,|P|)}
 +\frac{\sum_{(i,j)\in Q}[u_i^\top u_j]_+}{\max(1,|Q|)}\notag\\
 &\quad-\frac{0.5}{H}\sum_{r=1}^H H_b(\bar p_r),\label{eq:emb}
\end{align}
with $H_b(p)=-p\log(p+\epsilon)-(1-p)\log(1-p+\epsilon)$ and $\epsilon=10^{-8}$ as in the implementation. These terms encourage within-label similarity, penalize positive between-label similarity and discourage constant hidden activity; no generalization guarantee follows from them. Sigmoid and entropy occur only in the training loss, not the deployed circuit.

Forward activations are hard thresholds, with identity STE gradients. Adam uses a prespecified grid of $(\eta,\lambda)\in\{(0.02,0.0005),(0.005,0.005),(0.02,0.01)\}$, where $\lambda$ is coupled weight decay. Each candidate trains for at most 300 epochs. Batches are full when $|\mathcal T|\le1,024$ and deterministically shuffled groups of 512 otherwise, so ogbn-arxiv uses a batch-local pair regularizer. Gradients see only training labels; labels outside $\mathcal T\cup\mathcal V$ are replaced by $-1$ in the tensor supplied to training.

For candidate $h$, define validation accuracy and checkpoint selection by
\begin{align}
 A_{\mathcal V}(h,t)&=\frac1{|\mathcal V|}\sum_{i\in\mathcal V}
       \ind{\hat y_i^{\rm cur}(\phi_{h,t})=y_i},\notag\\
 t_h^*&=\min\arg\max_{1\le t\le T_h}A_{\mathcal V}(h,t),\notag\\
 h^*&=\arg\max_{h\in\mathcal H}A_{\mathcal V}(h,t_h^*).\label{eq:selection}
\end{align}
A checkpoint is saved on strict validation improvement, and training stops after 50 consecutive epochs without improvement or at epoch 300. Grid ties use the listed candidate order. The selected weights and calibrated thresholds are recorded with SHA-256 hashes before any final test evaluation. Test labels influence neither training, checkpointing, stopping nor candidate selection. A regression test rejects test-node assessment during selection and confirms that changing test labels leaves the selected and calibrated model identical.

\subsection{Validation-Only Output Calibration}
After restoring $\phi_{t^*}$, class firing thresholds are calibrated on $\mathcal V$ only. Coordinate ascent searches the current threshold, bounds 0.05 and 4, and midpoints between sorted distinct validation currents restricted to that range. Only strict improvements in validation one-hot correctness are accepted, with ties favoring the candidate closest to the current threshold; at most three passes are made. Weights and current predictions remain fixed. Calibration completes model selection; the calibrated checkpoint is frozen before test evaluation.
\begin{algorithm}[!t]
\caption{Output-Threshold Calibration}\label{alg:calib}
\begin{algorithmic}[1]
\REQUIRE validation currents $Z$, labels $y_{\mathcal V}$
\STATE initialize $\theta_{o,c}\leftarrow1$ for every class
\FOR{pass $=1$ to $3$}
\FOR{each class $c$}
\STATE form candidates: current threshold, 0.05, 4, and validation-current midpoints in $[0.05,4]$
\STATE select highest validation strict accuracy (ties: closest to current)
\STATE update $\theta_{o,c}$ only for a strict improvement
\ENDFOR
\ENDFOR
\RETURN $\theta_o$
\end{algorithmic}\end{algorithm}

\subsection{Spike-Native Realization}\label{sec:map}
\begin{figure*}[!t]\centering\includegraphics[width=\textwidth]{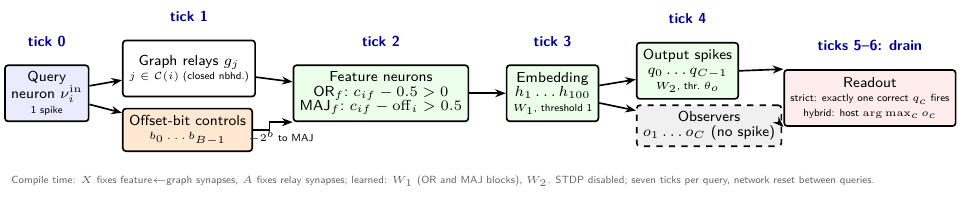}
\caption{Deployed DeNA-SGN circuit and timing. A single query spike at tick 0 reaches the closed neighborhood and the offset-bit controls at tick 1; feature neurons fire at tick 2, hidden units at tick 3, and output and observer neurons receive $W_2h$ at tick 4. Seven ticks are simulated and the network is reset between queries. Observer neurons never spike and allow a host to read class currents.}\label{fig:arch}\end{figure*}
Fig.~\ref{fig:arch} shows the circuit and Table~\ref{tab:pop} lists its neuron populations. Each node has a query neuron and a graph relay. The query neuron of node $i$ drives the relays of its closed neighborhood, and the relays drive the feature neurons through the fixed binary feature matrix, with one synapse for each non-zero entry. The classifier weights $W_1$ and $W_2$ are the learned synapses.
\begin{table}[!t]\centering\footnotesize\setlength{\tabcolsep}{3pt}
\caption{Neuron Populations of the Deployed Circuit}\label{tab:pop}
\begin{tabular}{@{}>{\raggedright\arraybackslash}p{1.45cm}>{\raggedright\arraybackslash}p{0.95cm}>{\raggedright\arraybackslash}p{1.15cm}>{\raggedright\arraybackslash}p{3.45cm}@{}}\toprule
Population & Count & Fires at tick & Role \\\midrule
Query & $N$ & 0 (input) & receives the single input spike \\
Relay & $N$ & 1 & forwards the query to its closed neighborhood \\
Control & $B$ & 1 & delivers $-2^b$ to Majority neurons if bit $b$ of $\off_i$ is set \\
OR feature & $D_{\mathrm{OR}}$ & 2 & fires if $c_{if}>0.5$ \\
Majority feature & $D_{\mathrm M}$ & 2 & fires if $c_{if}-\off_i>0.5$ \\
Hidden & $H$ & 3 & fires if $W_1x_i\ge\theta_h$ \\
Output & $C$ & 4 & fires if $z_c\ge\theta_{o,c}$ \\
Observer & $C$ & never & accumulates $z_c$, read at tick 5 \\\bottomrule
\end{tabular}\end{table}

\begin{figure}[!t]\centering\includegraphics[width=0.96\columnwidth]{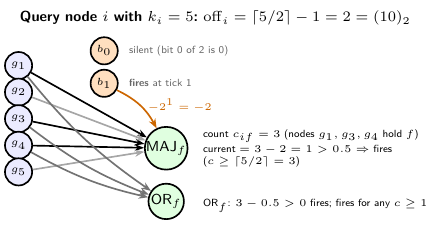}
\caption{Majority threshold by binary inhibition for $k_i=5$. The offset $2=(10)_2$ activates only control bit $b_1$, which subtracts 2 from every Majority neuron; a feature held by three of five nodes then reaches current $1>0.5$ and fires.}\label{fig:majority}\end{figure}

\emph{Degree-adaptive Majority threshold.} The effective Majority count threshold depends on query degree, but is realized using ordinary fixed-threshold neurons and signed control currents. Define the offset $\off_i=\lceil k_i/2\rceil-1$. For integer counts,
\begin{equation}
c-\off_i>0.5\iff c\ge\lceil k_i/2\rceil,\label{eq:majthr}
\end{equation}
so a Majority neuron with a fixed threshold of $0.5$ computes $m_{if}$ provided it receives $-\off_i$ from inhibitory controls. Writing $\off_i=\sum_b 2^b\beta_{ib}$ with $\beta_{ib}\in\{0,1\}$, control neuron $b$ spikes at tick 1 if and only if $\beta_{ib}=1$, because the query synapse to control $b$ exists only for set bits, and sends weight $-2^b$ to every Majority neuron (Fig.~\ref{fig:majority}). All queries share the same $B$ controls, so the number of control neurons is logarithmic in maximum closed-neighborhood size and independent of feature dimension. Their inhibitory connectivity still requires $B D_{\mathrm M}$ synapses.

\emph{Readout.} Output neurons spike when $z_c\ge\theta_{o,c}$, and the strict readout requires exactly the correct neuron to fire. Because an output spike resets its membrane, one non-firing \emph{observer} neuron per class accumulates $z_c$, and a host takes the argmax of the observer currents captured at the start of tick 5. This hybrid current readout is therefore not spike-only; the observers and their duplicated $W_2$ synapses are included in every resource figure.

\subsection{Circuit Accounting}\label{sec:count}
Let $D_{\mathrm{OR}}$ and $D_{\mathrm M}$ be the numbers of retained feature channels (exact-zero outgoing columns are pruned together with their incoming connections), $B$ the control bits, $H$ the hidden units, $C$ the classes and $A$ the closed adjacency. Here $X_{\mathrm{OR}}$ and $X_{\mathrm M}$ denote columns of the binary node-feature matrix retained for the corresponding channel; $\mathrm{nnz}$ counts nonzero entries and $\mathrm{popcnt}$ counts set bits. Then
\begin{align}
\#\text{neurons}&=2N+D_{\mathrm{OR}}+D_{\mathrm M}+B+H+2C,\label{eq:nn}\\
\#\text{synapses}&=\mathrm{nnz}(A)+\mathrm{nnz}(X_{\mathrm{OR}})+\mathrm{nnz}(X_{\mathrm M})\notag\\
&\quad+\textstyle\sum_i\mathrm{popcnt}(\off_i)+B D_{\mathrm M}\notag\\
&\quad+\mathrm{nnz}(W_1)+2\,\mathrm{nnz}(W_2).\label{eq:ns}
\end{align}
Since $\sum_i\mathrm{popcnt}(\off_i)\le NB$, a coarse synapse bound is $O(\mathrm{nnz}(A)+\mathrm{nnz}(X_{\mathrm{OR}})+\mathrm{nnz}(X_{\mathrm M})+(N+D_{\mathrm M})\log\Delta+dH+HC)$. The four-tick output arrival is logical simulator depth under ideal parallel fan-in and fan-out, not measured hardware latency. Weight sharing in the learned classifier does not remove physical connections, so each shared weight is counted wherever it is instantiated. The feature connectivity is compiled from the fixed binary matrix, so a change in graph or features requires recompilation.

\subsection{Validation Protocol}\label{sec:verify}
Simulating millions of idle connections for every query would be wasteful, so the harness retains for each query only its query neuron, its relays and all downstream feature, control and classifier neurons. The removed neurons are silent because they receive no input and have positive thresholds, so deleting them cannot change the remaining currents. Slices are grouped by degree bucket to bound padding, and several run concurrently in disjoint replicas, limited to 64 replicas, four million static classifier connections and 250,000 replica neurons per batch. All graph nodes remain represented in full-deployment accounting; no dataset is replaced by a node-sampled subgraph. Only designated test nodes are scored, rather than training/validation/unused-label nodes. Every test node is verified through native causal slices, not every query through the entire full network. The bulk compiler used for batching was tested against the original neuron-by-neuron construction, including full spike trains and event counts. Because slicing is an optimization rather than part of the model, the full deployment of every dataset is also built natively and queries spanning the test-index range and the minimum and maximum neighborhood size are replayed against it (Table~\ref{tab:coverage}). The isolation regression test additionally checks that test labels cannot affect model selection. Independent simultaneous queries use disjoint circuit replicas; multiplexed queries sharing the same feature/control bank are not verified here.

\emph{Precision contract.} Training uses float32, and the export runs the same stored weights in float64 to match the simulator's float64 accumulation. All retained OR/Majority feature, hidden and output bits and the class decision are asserted equal on every test query. Currents need not be bit-identical because reduction order can differ, so the largest current difference is recorded; it was zero in every case. Disagreements between the float32 training path and the export are counted separately and were zero for both classifications and spike patterns.

\subsection{Analytical Energy Model}
Following the common accumulate-versus-multiply-accumulate convention~\cite{zhu2022spikinggcn}, we assume 4.6~pJ per multiply-accumulate (MAC), 0.9~pJ per accumulate, 0.1~pJ per routed spike and 0.1~pJ per neuron update per tick, and we set weight reads and static power to zero. These are assumed operation costs, not measurements calibrated for float64 SuperNeuroMAT execution or a specified device. For $M$ learned MACs and $G$ neighborhood-gather additions per prediction, $\varepsilon$ synaptic events, $s$ spikes and $N_{\mathrm{dep}}$ deployed neurons clocked for seven ticks,
\begin{align}
E_{\mathrm{ANN}}&=4.6M+0.9G,\notag\\ E_{\mathrm{SNN}}&=0.9\varepsilon+0.1s+0.7N_{\mathrm{dep}}\quad[\text{pJ}].\label{eq:energy}
\end{align}
The learned dense ANN comparator uses $2dH+HC$ MACs per prediction. The SNN term counts every recorded signed synaptic arrival as one event, plus seven clocked updates of every deployed neuron. Sensitivities include cached ANN aggregation, zero-cost SNN neuron updates and a longer assumed tick duration. An optimized binary/sparse ANN could also use accumulates and skip zeros; these alternatives are not measured here. Binary comparisons, memory, physical interconnect beyond the per-spike proxy, placement, training, host injection and host argmax are excluded. Energy and average power are analytical estimates, not device measurements.

\section{Experimental Setup}\label{sec:setup}
\subsection{Datasets and Splits}
\begin{table*}[!t]\centering\footnotesize\setlength{\tabcolsep}{4pt}
\caption{Dataset Provenance and Loaded Graph Dimensions. Edges Count Unique Undirected Connections After Deduplication and Symmetrization, Including Any Source Self-Loops; Closed-Neighborhood Self-Inclusion Is Added During Compilation. $d$: Input Feature Dimension; $C$: Classes.}\label{tab:datasetsources}
\input{tables/t_dataset_provenance.tex}\end{table*}
\begin{table*}[!t]\centering\footnotesize\setlength{\tabcolsep}{4pt}
\caption{Exact Per-Run Split Counts. P: Provided Benchmark Masks; D: Derived KarateClub Split; F: Prespecified Amazon Photo Confirmatory Split. All Nodes and Edges Remain Visible as Transductive Context. Unused Labels Are Not Scored.}\label{tab:data}
\input{tables/t_splits.tex}\end{table*}
Table~\ref{tab:datasetsources} identifies each dataset and its benchmark source. Cora and CiteSeer follow the citation-network collection~\cite{sen2008collective}, and PubMed follows the diabetes citation network of Namata \emph{et al.}~\cite{namata2012query}; all three use the Planetoid benchmark release and its public masks~\cite{yang2016planetoid}. Wiki-CS supplies Wikipedia articles, hyperlink relations, word-embedding features and 20 official masks~\cite{mernyei2020wikics}. The official ogbn-arxiv benchmark provides title/abstract embeddings and publication-year splits~\cite{hu2020ogb,ogbarxivdata}. KarateClub uses Zachary's social graph~\cite{zachary1977karate} with the four modularity-derived classes, identity features and one labeled training node per class provided by PyG~\cite{pygkaratedata}; the first remaining node per class supplies validation and the rest test. This is supervised node classification, not unsupervised community detection. Wiki-CS uses all 20 supplied training/validation masks and their shared test set. Its supplied stopping masks are not used; early stopping uses the reported validation set.

Amazon Photo, the product co-purchase benchmark with review bag-of-words features~\cite{shchur2018pitfalls}, is a new confirmatory benchmark added after architectural development on the original six datasets. Before evaluating its test scores, we fixed a stratified split using seed 20261020: 20 training and 30 validation nodes per class, with all remaining nodes assigned to testing. The same split is used for three initialization seeds. Binary features use positive presence for citation datasets and Amazon Photo, identity features for KarateClub, and training-node feature medians for Wiki-CS and ogbn-arxiv. Neighborhood aggregation uses the complete symmetrized graph with self-inclusion. No node-sampled graph substitutes for a dataset. This transductive, symmetrized arxiv protocol is not its directed leaderboard protocol.

The original six datasets were examined during earlier architecture development. Their new results are validation-selected replications; Amazon Photo supplies the additional dataset whose test outcomes were unexamined when this study's protocol was frozen. Historical exploratory runs are preserved separately and do not supply the primary tables below.

\subsection{Training, Controls and Protected Evaluation}
The hidden width is 100. The three optimizer candidates, maximum epoch budget, validation-only selection rule and calibration are specified in Section~\ref{sec:method}. Every selected model across the study is locked before final evaluation. The run bundles record the protocol, split IDs, source snapshots, per-epoch validation logs, candidate scores, selected-model hashes and final predictions.

Cora OR controls use exactly the same splits, seeds, optimizer grid, initialization rule, objective and stopping budget as OR+Majority; their majority inputs are clamped to zero and the corresponding circuit connections are pruned. A standard ANN GAT reference uses eight heads of eight hidden channels, ELU, dropout 0.6, Adam with learning rate 0.005 and weight decay 0.0005, maximum 300 epochs and validation patience 50. It uses the same Cora split IDs and initialization seeds. GAT is an ANN accuracy reference, with no SNN mapping claim.

Experiments use SuperNeuroMAT 3.5.0 in sparse CPU mode, PyTorch 2.14.1 and PyTorch Geometric 2.8.1~\cite{fey2019pyg}. All test queries for every selected spiking model execute through the native simulator. Full-deployment resource counts include all graph nodes and supporting neurons; selected whole-network replays additionally validate the causal-slice implementation.

\subsection{Metrics and Meaning of Variability}
Current accuracy uses the host argmax of simulator-observed currents and is checked against the ANN export. Strict spike accuracy requires the entire output vector to equal the correct one-hot label. We also report silent/multiple-output rates, macro-F1, balanced accuracy, negative log-likelihood, Brier score and expected calibration error. The latter three use a host softmax as a diagnostic, not a circuit operation.

For run accuracies $a_1,\ldots,a_n$, we report
\begin{equation}
 \bar a=\frac1n\sum_{r=1}^n a_r,\qquad
 s=\sqrt{\frac{\sum_{r=1}^n(a_r-\bar a)^2}{n-1}}.\label{eq:sd}
\end{equation}
Thus $\bar a\pm s$ describes sample standard deviation across runs, not a confidence interval or an observed minimum--maximum interval. 
Fixed splits use initialization seeds 42, 43 and 44. Wiki-CS public uses 20 correlated masks with cyclic seeds, so its variability reflects both mask and initialization. These are not 20 independent datasets or independent test samples.


\section{Results}\label{sec:results}
The evaluation establishes three main findings. \textbf{Useful graph classification:} the new confirmatory Amazon Photo experiment reaches \PhotoMean$\pm$\PhotoSD\% current-readout accuracy, while Wiki-CS reaches \WikiPublicMean$\pm$\WikiPublicSD\% over 20 public masks. \textbf{Decision-matched execution:} all \AllNativeQueries{} native test-query executions preserve the exported model's class and output-spike decisions. \textbf{An effective extra graph channel:} the matched Cora comparison records a mean gain of \MajorityGain{} points from adding Majority on the public split, with \MajorityExtraNeurons{} additional deployed neurons. The following subsections separate predictive accuracy, mapping fidelity and implementation cost.

\subsection{Held-Out Test Accuracy and Native Mapping}
\begin{table}[!t]\centering\footnotesize\setlength{\tabcolsep}{3pt}
\caption{Validation-Selected Final Test Results (Mean $\pm$ Sample SD). P: Provided Benchmark Masks; D: Derived KarateClub Split; F: Prespecified Amazon Photo Split. Current Accuracy Is Identical for the ANN Export and Native SNN With Host Argmax; Strict Accuracy Requires the Correct One-Hot Spike Vector.}\label{tab:main}
\input{tables/t_main.tex}\end{table}
Table~\ref{tab:main} reports the final test evaluation of \MainRuns{} locked OR+Majority models. Public current-readout accuracy is \CoraMean$\pm$\CoraSD\% on Cora, \CiteSeerMean$\pm$\CiteSeerSD\% on CiteSeer and \PubMedMean$\pm$\PubMedSD\% on PubMed. On the new Amazon Photo confirmatory dataset, the corresponding result is \PhotoMean$\pm$\PhotoSD\%, with strict spike accuracy \PhotoStrictMean$\pm$\PhotoStrictSD\%. The three individual Photo current scores are \PhotoRunValues\%. The protocol and all candidate choices were fixed before these test predictions were assessed.

Every retained OR/Majority feature, hidden bit, output bit and current-argmax class decision matches the float64 ANN export on all \MainQueries{} benchmark test-query executions. These are repeated executions across runs, not a count of distinct graph nodes. Three additional OR controls contribute \ControlQueries{} native test-query executions with the same decision checks. The precision diagnostics separately count differences from the float32 training path. The maximum native-versus-export current difference is \MaxCurrentError{}; this is empirical finite-precision verification of evaluated queries, not a universal floating-point equivalence theorem.
\begin{table}[!t]\centering\footnotesize\setlength{\tabcolsep}{3pt}
\caption{Native Verification Coverage. Every Test Query Uses Native Causal Slices. Full Replays Are Selected Queries on One Entire Native Deployment per Dataset. OR Controls Are Additional to This Table.}\label{tab:coverage}
\input{tables/t_coverage.tex}\end{table}
The seven full native deployments reproduce the counted neuron/synapse totals and the sliced spike/event activity on \FullReplays{} selected queries spanning test-index and neighborhood-size extremes. Full-network simulation for every test query is not claimed.
\begin{figure*}[!t]\centering\includegraphics[width=\textwidth]{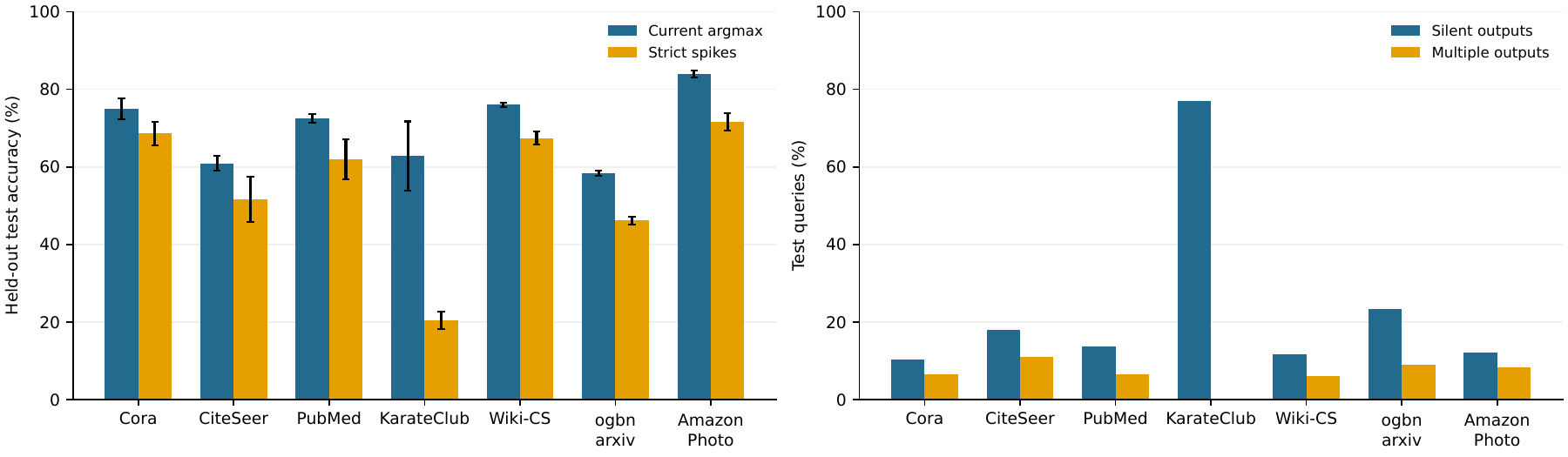}
\caption{Public-protocol results and the new Amazon Photo confirmatory result. Left: final test accuracy after validation-only selection (mean $\pm$ sample SD). Right: silent and multiple-output frequencies. All percentage axes start at zero.}\label{fig:results}\end{figure*}

\subsection{Decoding and Diagnostic Metrics}\label{sec:decode}
\begin{table}[!t]\centering\footnotesize\setlength{\tabcolsep}{3pt}
\caption{Decoding Rules (Percent), Public Protocol and Amazon Photo. Binary Argmax Selects the First Maximal Output Bit, Including Class 0 for Silence.}\label{tab:decode}
\input{tables/t_decode.tex}\end{table}
Current readout and strict spike correctness measure different behaviors. Current argmax can use subthreshold evidence, while strict correctness requires exactly one correct firing output. Class-specific calibration means strict-correct/current-wrong cases can also occur, so silence and multiplicity alone do not decompose the difference between accuracies. A winner-take-all extension cannot guarantee the correct class and would require its own complete circuit and verification. Supplementary Table~S1 reports balanced accuracy, macro-F1 and calibration diagnostics; the per-run evidence records confusion matrices and per-class precision, recall and F1.

\subsection{Matched OR Ablation and GAT Reference}\label{sec:ablate}
\begin{table*}[!t]\centering\footnotesize\setlength{\tabcolsep}{5pt}
\caption{Matched Cora Evaluation. OR and OR+Majority Share Splits, Seeds, Candidate Grid and Training Budget, and Both Receive Native SNN Verification. GAT Is an ANN Reference With Validation-Selected Checkpoints. Resource Counts Are Full-Deployment Run Means.}\label{tab:abl}
\input{tables/t_ablation_validation.tex}\end{table*}
\begin{figure}[!t]\centering\includegraphics[width=\columnwidth]{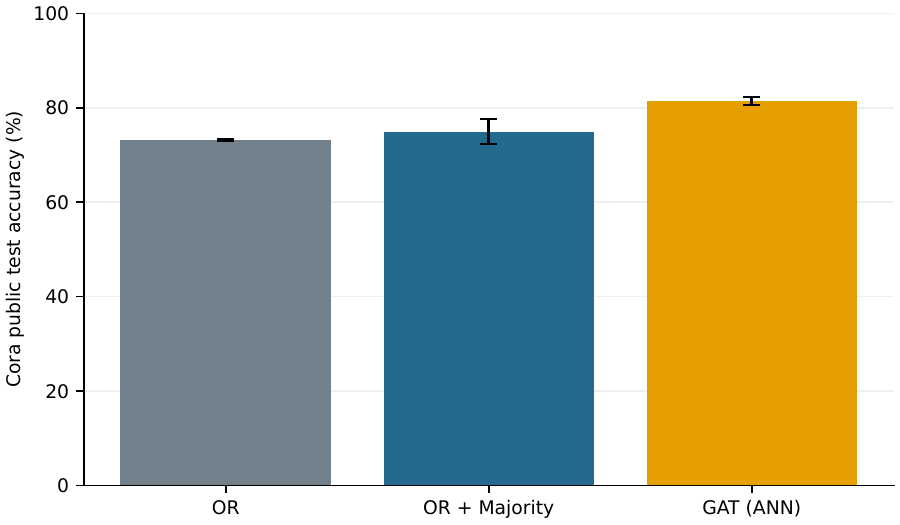}
\caption{Cora public test accuracy after validation-only selection. Error bars are sample SD over three seeds; the accuracy axis starts at zero.}\label{fig:pareto}\end{figure}
The matched public-split comparison yields \ORMean$\pm$\ORSD\% for OR and \CoraMean$\pm$\CoraSD\% for OR+Majority, a mean difference of \MajorityGain{} percentage points. The ANN GAT reference reaches \GATMean$\pm$\GATSD\%. The extra channel requires \MajorityExtraNeurons{} additional neurons (\MajorityNeuronIncrease\% over OR). These matched comparisons show the value of retaining a second level of neighborhood support at the reported operating point. The gains are run means over three seeds; statistical significance and universal superiority are not inferred. Earlier architectural investigations motivated the design, but their exploratory scores are not pooled with this comparison.

\subsection{Comparison With Reported Spiking Implementations}\label{sec:compare}
\begin{table*}[!t]\centering\footnotesize\setlength{\tabcolsep}{3pt}
\caption{Reported Citation-Node Classification Accuracy (\%). Spiking Scope Distinguishes Spike-Only Inference, a Spike-Native Core With External Readout, and Hybrid Graph Processing. Original Evaluation Contexts Are Preserved. External $\pm$ Values Follow Their Sources; DeNA-SGN Uses Sample SD Across Three Seeds.}\label{tab:published}
\input{tables/t_published_comparison.tex}
\par\vspace{2pt}\begin{minipage}{0.98\textwidth}\scriptsize
\textbf{Purely spiking?} Yes: online aggregation, classifier propagation and the output decision use spikes; offline training/feature encoding are excluded. Core only: spike-native dynamics with external current or synaptic-weight decoding. No: real-valued graph convolution precedes spike encoding. This column does not certify physical hardware execution. $\dagger$ ICONS 2022 marks PubMed 54\% as preliminary, based on 6,300 evaluated test samples at submission; its Table 1 separately lists 1,000 test nodes. $\ddagger$ The Lava result is validation accuracy on 140 Cora nodes, not held-out test accuracy. ICMLA 2023 Table III explicitly excludes node features; its Table I lists 140 CiteSeer training nodes, versus 120 here. Exact masks are not established across papers. Dashes mean not reported in the cited result.\end{minipage}
\end{table*}
The graph aggregation and threshold classifier of DeNA-SGN execute through spike-driven neuron/synapse dynamics. Its strict output-spike row also obtains the decision from emitted class spikes; silence and multiple firing are scored as errors. The higher-accuracy current-readout row uses host argmax and is therefore marked ``Core only,'' as are STDP systems that decode learned synaptic weights. SpikingGCN variants are marked ``No'' because their graph convolution is real-valued before spike encoding. This separates the closest spike-native-core comparisons from hybrid references without conflating execution scope with accuracy.

Table~\ref{tab:published} places the NEST-based studies of Cong \emph{et al.} alongside related spiking graph systems and the present model. ICONS 2022 reports NEST graph-only averages of 56\%, 41\% and 54\% over five random-split runs~\cite{cong2022semisupervised}. ICMLA 2023 improves its optimized structure-only scores to 67\%, 51\% and 79\% using LavaBO~\cite{cong2023hpofeat}; those numbers are separate from its feature-inclusion experiments. LavaBO names the optimization framework and does not by itself establish Loihi execution. Snyder \emph{et al.}'s fixed-point Lava result is labeled as validation accuracy in the table~\cite{snyder2024transductive}.

For the spike-only DeNA-SGN decision rule, public accuracies are \CoraStrictMean$\pm$\CoraStrictSD\% on Cora, \CiteSeerStrictMean$\pm$\CiteSeerStrictSD\% on CiteSeer and \PubMedStrictMean$\pm$\PubMedStrictSD\% on PubMed. These are the appropriate DeNA-SGN values for a strict spike-output comparison; the host-current row is a separately labeled readout option. Cong \emph{et al.}'s structure-only results remain relevant spike-native-core references, but node features, masks, optimization and readout still differ. Thus grouping by spiking scope improves the comparison but does not make cross-paper scores a controlled head-to-head experiment. The matched OR/OR+Majority comparison in Table~\ref{tab:abl} supplies that within-study evidence.
\begin{figure*}[!t]\centering\includegraphics[width=\textwidth]{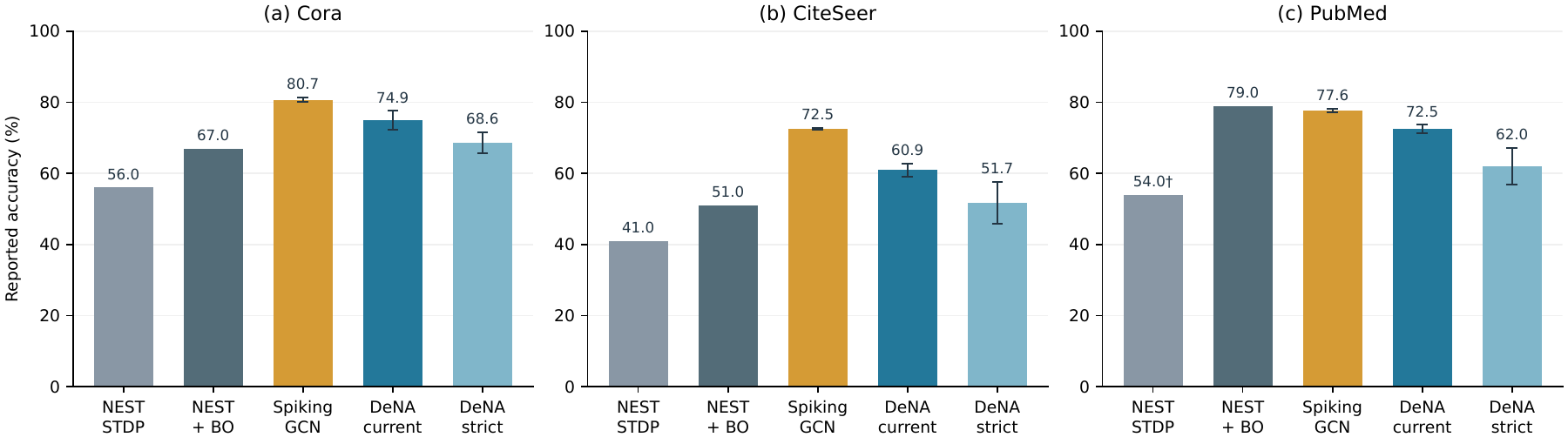}
\caption{Published citation benchmarks and the present validation-selected results. Axes start at zero; all panels use the same 0--100\% scale. The NEST STDP and NEST+BO bars correspond to the ICONS 2022 and ICMLA 2023 studies; SpikingGCN uses Split I~\cite{cong2022semisupervised,cong2023hpofeat,zhu2022spikinggcn}. Error bars are shown only when variability is reported and retain each study's definition. $\dagger$ Preliminary PubMed coverage is explained in Table~\ref{tab:published}. Feature use and protocols differ, so the plot is descriptive.}\label{fig:published}\end{figure*}

\subsection{Complete Circuit and Analytical Cost}\label{sec:cost}
\begin{table*}[!t]\centering\footnotesize\setlength{\tabcolsep}{5pt}
\caption{Full-Deployment Resources and Modeled Cost per Prediction, Public Protocol and Amazon Photo. Spikes/Events Include Graph, Control and Readout Activity; Ratio Is ANN/SNN Under Eq.~\eqref{eq:energy}. Values Are Run Means.}\label{tab:cost}
\input{tables/t_cost.tex}\end{table*}
Table~\ref{tab:cost} includes query neurons, graph relays, retained OR/Majority channels, inhibitory controls, classifier neurons and nonfiring current observers. No gating or synchronization circuitry is excluded from these counts. Per-prediction spikes and signed synaptic arrivals are recorded from native execution, with mean, standard deviation, minimum, median, 95th percentile and maximum retained in each run report. Logical learned-weight parameters and physically retained nonzero synapses are distinct quantities and are recorded separately.
\begin{figure*}[!t]\centering\includegraphics[width=\textwidth]{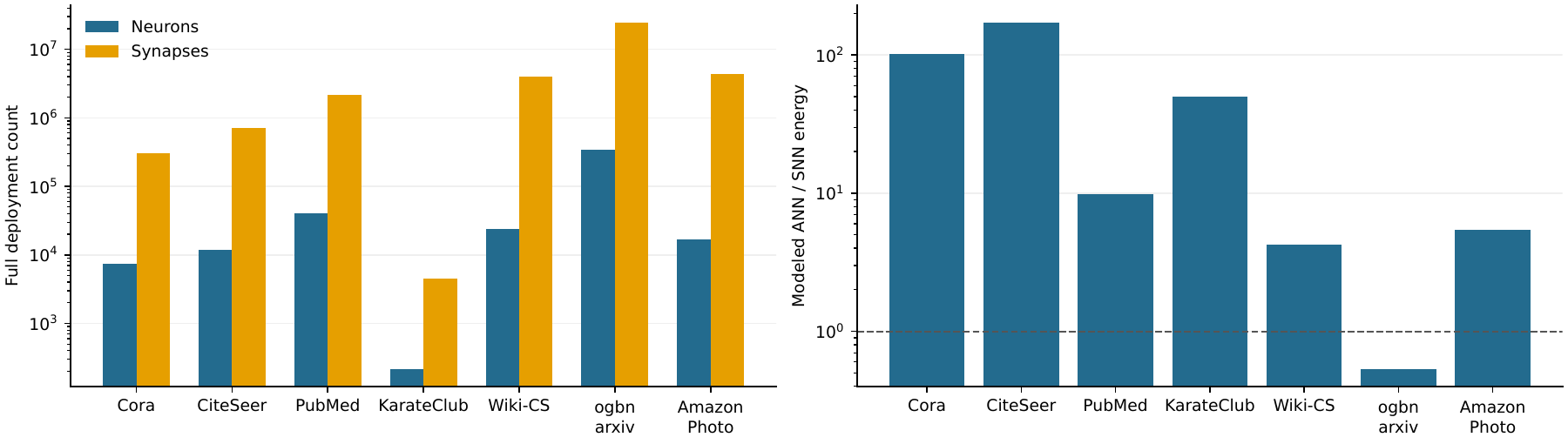}
\caption{Left: complete deployed neuron and synapse counts. Right: modeled ANN-to-SNN energy ratio; the dashed line marks equal modeled energy. These are operation-cost estimates, not device measurements.}\label{fig:cost}\end{figure*}
\emph{A resource-accounting finding.} Under the stated dense-MAC comparator, the nominal modeled ANN/SNN energy ratios are \CoraEnergyRatio$\times$ on Cora and \CiteSeerEnergyRatio$\times$ on CiteSeer. The ratio becomes \ArxivEnergyRatio$\times$ on ogbn-arxiv, where clocking the much larger deployed neuron population dominates the modeled SNN cost. This crossover is useful design evidence: sparse spike traffic alone does not ensure low cost when inactive neurons still incur clocked updates. These are conditional operation-cost estimates, rather than measured device savings.

The energy comparison charges dense learned MACs to the ANN and recorded arrivals/spikes plus all deployed neuron updates to the SNN. Consequently the ratio depends on graph size, activity and the comparator. Clocking every idle neuron penalizes large deployments. The reports retain cached-ANN and zero-neuron-update sensitivities and modeled average power for explicitly assumed tick durations; these quantities support reproducible accounting rather than measured hardware-efficiency claims.


\section{Discussion}\label{sec:disc}
\subsection{What the Findings Establish}
The strongest result is the combination of predictive utility and verified implementation: \PhotoMean\% current accuracy on the new confirmatory graph, native agreement across \AllNativeQueries{} test-query executions, and complete circuit accounting up to the 169,343-node arxiv graph. Majority adds useful discrimination beyond OR by distinguishing isolated support from broad neighborhood support; the matched Cora public-split comparison quantifies its benefit under the same masks, seeds and training budget. These findings support deterministic threshold aggregation as a practical, inspectable operating point for spiking graph classification.

\subsection{Implications for Model and Circuit Design} The ANN is trained with hard thresholds from the outset, avoiding activation-rescaling and rate-coding approximations in the mapped classifier. Ideal logical equivalence follows from the construction; finite-precision execution is checked empirically. The resulting model is less expressive than learned continuous attention, but exposes every graph, control and classifier operation for circuit compilation.

\subsection{Deployment and Evaluation Scope} The logical depth and explicit connectivity provide a concrete design for event-driven implementation. Device-specific quantization, fan-in limits, placement, routing and physical timing remain to be evaluated. Host current argmax and offline feature encoding are explicit external operations; strict spike-only results quantify the existing alternative. Power and energy are analytical and exclude memory and physical interconnect costs beyond a fixed spike proxy. Thus the study establishes simulator-matched execution and complete accounting, while leaving physical efficiency to measured device implementations.

\noindent \paragraph{Generalization} Test labels are excluded from the new training/selection pipeline. The original six benchmarks nevertheless informed earlier architecture development; the added Amazon Photo protocol supplies an additional confirmatory evaluation. Three-run standard deviations are descriptive, and Wiki-CS masks share test nodes. Binarization discards continuous magnitudes, arxiv is symmetrized and changes to graph/features can require recompilation. These scope conditions do not alter the verified circuit-equivalence result.

\section{Conclusion}\label{sec:concl}
DeNA-SGN demonstrates that useful graph classification can be paired with a constructive, decision-matched spiking realization. Its OR and degree-adaptive Majority channels drive an offline-trained threshold classifier whose graph, control and readout circuits are explicitly compiled and counted. The new Amazon Photo experiment achieves \PhotoMean$\pm$\PhotoSD\% current-readout accuracy and \PhotoStrictMean$\pm$\PhotoStrictSD\% strict spike accuracy; Wiki-CS achieves \WikiPublicMean$\pm$\WikiPublicSD\% over 20 masks.  Complete accounting also identifies an important deployment condition: activity sparsity must be considered together with the cost of inactive neuron updates. The resulting construction, verified execution and reproducible evidence provide a foundation for quantized mappings, improved spike-only readout and measured neuromorphic implementations.

\section*{Acknowledgment}
The authors thank the developers of the SuperNeuroMAT simulator for making it openly available. This research is sponsored by the Office of the Laboratory Director, Oak Ridge National Laboratory's Operational Excellence Initiatives, which is supported by the United
States Department of Energy (DOE)’s Office of Science under Contract No. DE-AC05-00OR22725. 

\bibliographystyle{IEEEtran}
\bibliography{ref}
\balance
\end{document}

%% file: tables/validation_numbers.tex
\newcommand{\CoraMean}{74.90}
\newcommand{\CoraSD}{2.67}
\newcommand{\CoraStrictMean}{68.57}
\newcommand{\CoraStrictSD}{2.94}
\newcommand{\CiteSeerMean}{60.93}
\newcommand{\CiteSeerSD}{1.90}
\newcommand{\CiteSeerStrictMean}{51.67}
\newcommand{\CiteSeerStrictSD}{5.90}
\newcommand{\PubMedMean}{72.53}
\newcommand{\PubMedSD}{1.12}
\newcommand{\PubMedStrictMean}{61.97}
\newcommand{\PubMedStrictSD}{5.20}
\newcommand{\PhotoMean}{83.93}
\newcommand{\PhotoSD}{0.93}
\newcommand{\PhotoStrictMean}{71.61}
\newcommand{\PhotoStrictSD}{2.22}

\newcommand{\MainRuns}{38}
\newcommand{\MainQueries}{293,577}
\newcommand{\FullReplays}{47}
\newcommand{\ControlQueries}{3,000}
\newcommand{\MaxCurrentError}{0}
\newcommand{\PhotoRunValues}{84.59, 82.87, 84.33}
\newcommand{\ORMean}{73.27}
\newcommand{\ORSD}{0.21}
\newcommand{\GATMean}{81.47}
\newcommand{\GATSD}{0.90}
\newcommand{\MajorityGain}{1.63}

%% file: tables/research_findings_numbers.tex
\newcommand{\AllMappedModels}{41}
\newcommand{\AllNativeQueries}{296,577}
\newcommand{\MajorityExtraNeurons}{551}
\newcommand{\MajorityNeuronIncrease}{7.91}
\newcommand{\WikiPublicMean}{75.99}
\newcommand{\WikiPublicSD}{0.54}
\newcommand{\CoraEnergyRatio}{102.35}
\newcommand{\CiteSeerEnergyRatio}{171.03}
\newcommand{\ArxivEnergyRatio}{0.53}

%% file: tables/t_dataset_provenance.tex
\begin{tabular}{@{}lrrrrl@{}}
\toprule
Dataset / source & Nodes & Undirected edges & $d$ & $C$ & Node representation before encoding \\
\midrule
Cora~\cite{sen2008collective,yang2016planetoid} & 2,708 & 5,278 & 1433 & 7 & Binary word presence \\
CiteSeer~\cite{sen2008collective,yang2016planetoid} & 3,327 & 4,552 & 3703 & 6 & Binary word presence \\
PubMed~\cite{namata2012query,yang2016planetoid} & 19,717 & 44,324 & 500 & 3 & TF--IDF word attributes \\
KarateClub~\cite{zachary1977karate,pygkaratedata} & 34 & 78 & 34 & 4 & Identity features (PyG) \\
Wiki-CS~\cite{mernyei2020wikics} & 11,701 & 216,123 & 300 & 10 & Article-word embeddings \\
ogbn-arxiv~\cite{hu2020ogb,ogbarxivdata} & 169,343 & 1,157,799 & 128 & 40 & Title/abstract embeddings \\
Amazon Photo~\cite{shchur2018pitfalls} & 7,650 & 119,081 & 745 & 8 & Review bag-of-words \\
\bottomrule
\end{tabular}

%% file: tables/t_splits.tex
\begin{tabular}{@{}lrrrrrr@{}}
\toprule
Dataset & Protocol & Runs & Train & Validation & Test & Unused \\
\midrule
Cora & provided & 3 & 140 & 500 & 1000 & 1068 \\
CiteSeer & provided & 3 & 120 & 500 & 1000 & 1707 \\
PubMed & provided & 3 & 60 & 500 & 1000 & 18157 \\
KarateClub & derived & 3 & 4 & 4 & 26 & 0 \\
Wiki-CS & provided & 20 & 580 & 1769 & 5847 & 3505 \\
ogbn-arxiv & provided & 3 & 90941 & 29799 & 48603 & 0 \\
AmazonPhoto & confirmatory & 3 & 160 & 240 & 7250 & 0 \\
\bottomrule
\end{tabular}

%% file: tables/t_main.tex
\begin{tabular}{@{}llccc@{}}
\toprule
Dataset & Prot. & Current (\%) & Strict (\%) & Macro-F1 \\
\midrule
Cora & P & 74.90$\pm$2.67 & 68.57$\pm$2.94 & 0.741 \\
CiteSeer & P & 60.93$\pm$1.90 & 51.67$\pm$5.90 & 0.568 \\
PubMed & P & 72.53$\pm$1.12 & 61.97$\pm$5.20 & 0.723 \\
KarateClub & D & 62.82$\pm$8.88 & 20.51$\pm$2.22 & 0.532 \\
Wiki-CS & P & 75.99$\pm$0.54 & 67.36$\pm$1.66 & 0.721 \\
ogbn-arxiv & P & 58.36$\pm$0.69 & 46.20$\pm$0.99 & 0.244 \\
AmazonPhoto & F & 83.93$\pm$0.93 & 71.61$\pm$2.22 & 0.812 \\
\bottomrule
\end{tabular}

%% file: tables/t_coverage.tex
\begin{tabular}{@{}lrrr@{}}
\toprule
Dataset & Runs & Native test queries & Full replays \\
\midrule
Cora & 3 & 3,000 & 7 \\
CiteSeer & 3 & 3,000 & 7 \\
PubMed & 3 & 3,000 & 7 \\
KarateClub & 3 & 78 & 5 \\
Wiki-CS & 20 & 116,940 & 7 \\
ogbn-arxiv & 3 & 145,809 & 7 \\
AmazonPhoto & 3 & 21,750 & 7 \\
\bottomrule
\end{tabular}

%% file: tables/t_decode.tex
\begin{tabular}{@{}lrrrrr@{}}
\toprule
Dataset & Current & Strict & Bin. Argmax & Silent & Multi \\
\midrule
Cora & 74.90 & 68.57 & 73.03 & 10.40 & 6.57 \\
CiteSeer & 60.93 & 51.67 & 57.57 & 17.90 & 11.07 \\
PubMed & 72.53 & 61.97 & 67.80 & 13.73 & 6.57 \\
KarateClub & 62.82 & 20.51 & 57.69 & 76.92 & 0.00 \\
Wiki-CS & 75.99 & 67.36 & 70.44 & 11.80 & 6.07 \\
ogbn-arxiv & 58.36 & 46.20 & 49.89 & 23.46 & 8.96 \\
AmazonPhoto & 83.93 & 71.61 & 75.90 & 12.22 & 8.34 \\
\bottomrule
\end{tabular}

%% file: tables/t_ablation_validation.tex
\begin{tabular}{@{}lrrrr@{}}
\toprule
Model & Current (\%) & Strict (\%) & Neurons & Synapses \\
\midrule
OR & 73.27$\pm$0.21 & 66.03$\pm$5.13 & 6,970 & 209,749 \\
OR+Majority & 74.90$\pm$2.67 & 68.57$\pm$2.94 & 7,521 & 304,868 \\
GAT & 81.47$\pm$0.90 & -- & -- & -- \\
\bottomrule
\end{tabular}

%% file: tables/t_published_comparison.tex
\begin{tabular}{@{}>{\raggedright\arraybackslash}p{3.9cm}>{\raggedright\arraybackslash}p{1.3cm}c>{\raggedright\arraybackslash}p{3.6cm}rrr@{}}
\toprule
Method & Purely spiking? & Features & Reported evaluation / readout & Cora & CiteSeer & PubMed \\
\midrule
Cong et al., ICONS 2022~\cite{cong2022semisupervised} & Core only & no & Five random-split runs; Table 2 & 56 & 41 & 54$^{\dagger}$ \\
Cong et al., ICMLA 2023 (BO)~\cite{cong2023hpofeat} & Core only & no & Optimized structure-only; Table III & 67 & 51 & 79 \\
Fixed-point Lava~\cite{snyder2024transductive} & Core only & no & Cora validation; 140 nodes & 62.86$^{\ddagger}$ & -- & -- \\
Structure-only SuperNeuroMAT~\cite{date2026superneuromat} & Core only & no & Citation example; Section 6.2.2 & 62.7 & 42.4 & 54.7 \\
\textbf{DeNA-SGN: strict spikes} & Yes & yes & Public test; spike decision & \CoraStrictMean$\pm$\CoraStrictSD & \CiteSeerStrictMean$\pm$\CiteSeerStrictSD & \PubMedStrictMean$\pm$\PubMedStrictSD \\
\textbf{DeNA-SGN: current readout} & Core only & yes & Public test; host argmax & \CoraMean$\pm$\CoraSD & \CiteSeerMean$\pm$\CiteSeerSD & \PubMedMean$\pm$\PubMedSD \\
SpikingGCN~\cite{zhu2022spikinggcn} & No & yes & Split I test; 10 runs & 80.7$\pm$0.6 & 72.5$\pm$0.2 & 77.6$\pm$0.5 \\
SpikingGCN-N~\cite{zhu2022spikinggcn} & No & yes & Split I test; 10 runs & 81$\pm$0.4 & 72.9$\pm$0.1 & 78.5$\pm$0.2 \\
\bottomrule
\end{tabular}

%% file: tables/t_cost.tex
\begin{tabular}{@{}lrrrrrrr@{}}
\toprule
Dataset & Neurons & Synapses & Spikes & Events & ANN (nJ) & SNN (nJ) & Ratio \\
\midrule
Cora & 7,521 & 304,868 & 111.7 & 8,485 & 1321.7 & 12.91 & 102.35 \\
CiteSeer & 11,842 & 715,338 & 141.5 & 12,925 & 3409.6 & 19.94 & 171.03 \\
PubMed & 40,521 & 2,179,908 & 223.7 & 20,232 & 461.6 & 46.60 & 9.91 \\
KarateClub & 214 & 4,461 & 16.7 & 569 & 33.1 & 0.66 & 49.94 \\
Wiki-CS & 24,133 & 4,046,321 & 505.7 & 55,995 & 285.8 & 67.34 & 4.24 \\
ogbn-arxiv & 339,135 & 24,380,220 & 228.0 & 22,546 & 136.9 & 257.71 & 0.53 \\
AmazonPhoto & 16,916 & 4,380,229 & 1130.8 & 130,193 & 698.9 & 129.13 & 5.41 \\
\bottomrule
\end{tabular}